\documentclass[runningheads]{llncs}

\usepackage{eccv}

\usepackage{eccvabbrv}

\usepackage{graphicx}
\usepackage{booktabs}
\usepackage{amsmath}
\usepackage{multirow}
\usepackage{amssymb}
\usepackage{subcaption}
\usepackage{algorithm}
\usepackage{algorithmic}
\usepackage{makecell}
\usepackage{stfloats}
\usepackage{wrapfig}
\usepackage{threeparttable}
\usepackage{marvosym}
\usepackage[accsupp]{axessibility}  

\usepackage{colortbl}
\usepackage[table,xcdraw]{xcolor}
\definecolor{cellgreen}{HTML}{C5EFCE}
\definecolor{fontgreen}{HTML}{026102}
\definecolor{cellgray}{gray}{0.94}
\definecolor{fontgray}{gray}{0.50}

\usepackage{hyperref}

\usepackage{orcidlink}

\begin{document}

\title{Learning to Suppress SPAD-based LiDAR Flare} 

\titlerunning{Learning to Suppress SPAD-based LiDAR Flare}

\author{Xuanya Zhu\inst{1}\thanks{Work done during an internship at Advanced Technology Center, Sony (China) Ltd.},
Linghao Shen\inst{2}\textsuperscript{(\Letter)}}

\authorrunning{X.Zhu et al.}

\institute{Centre for Vision, Speech and Signal Processing, University of Surrey, UK
\and
Advanced Technology Center, Sony (China) Ltd., China\\
\email{marka.shen@sony.com}}

\maketitle

\begin{abstract}
  Single-Photon Avalanche Diode (SPAD)-based Light Detection and Ranging (LiDAR) is emerging for autonomous vehicles due to its high sensitivity and precise depth sensing capabilities. However, flare caused by excessive photon returns or pile-up effects can lead to incorrect depth estimation and exaggerated boundaries in point clouds, resulting in severe distortions of geometric measurements, making flare suppression essential for safety-critical applications.
  Existing flare mitigation methods primarily operate at the hardware or signal-processing levels. While effective under specific configurations, they are largely rule-based and configuration-dependent, lacking learnable representations that generalize across diverse sensing scenarios.
  In this work, we reformulate flare suppression as a semantic segmentation problem, enabling data-driven learning of geometric and photometric cues directly from SPAD measurements.
  We first benchmark representative segmentation models on the newly introduced SPAD flare dataset and observe that they struggle to exploit the intrinsic multi-echo characteristics of SPAD signals. Motivated by this observation, we propose \textbf{Physically-Informed segmentation for LiDAR Flare (PILF)}, a learning-based approach that treats the first and second echoes, together with ambient illumination, as distinct modalities, aggregating cross-echo information while jointly encoding geometric and photometric features.
  Experiments across multiple real-world scenes demonstrate that PILF significantly outperforms compared segmentation models, achieving up to \textbf{79.32\% mIoU}, and providing an effective solution for SPAD-based LiDAR flare suppression.
\end{abstract}

\section{Introduction}
\begin{figure*}[t]
    \centering
    \includegraphics[width=1.0\linewidth]{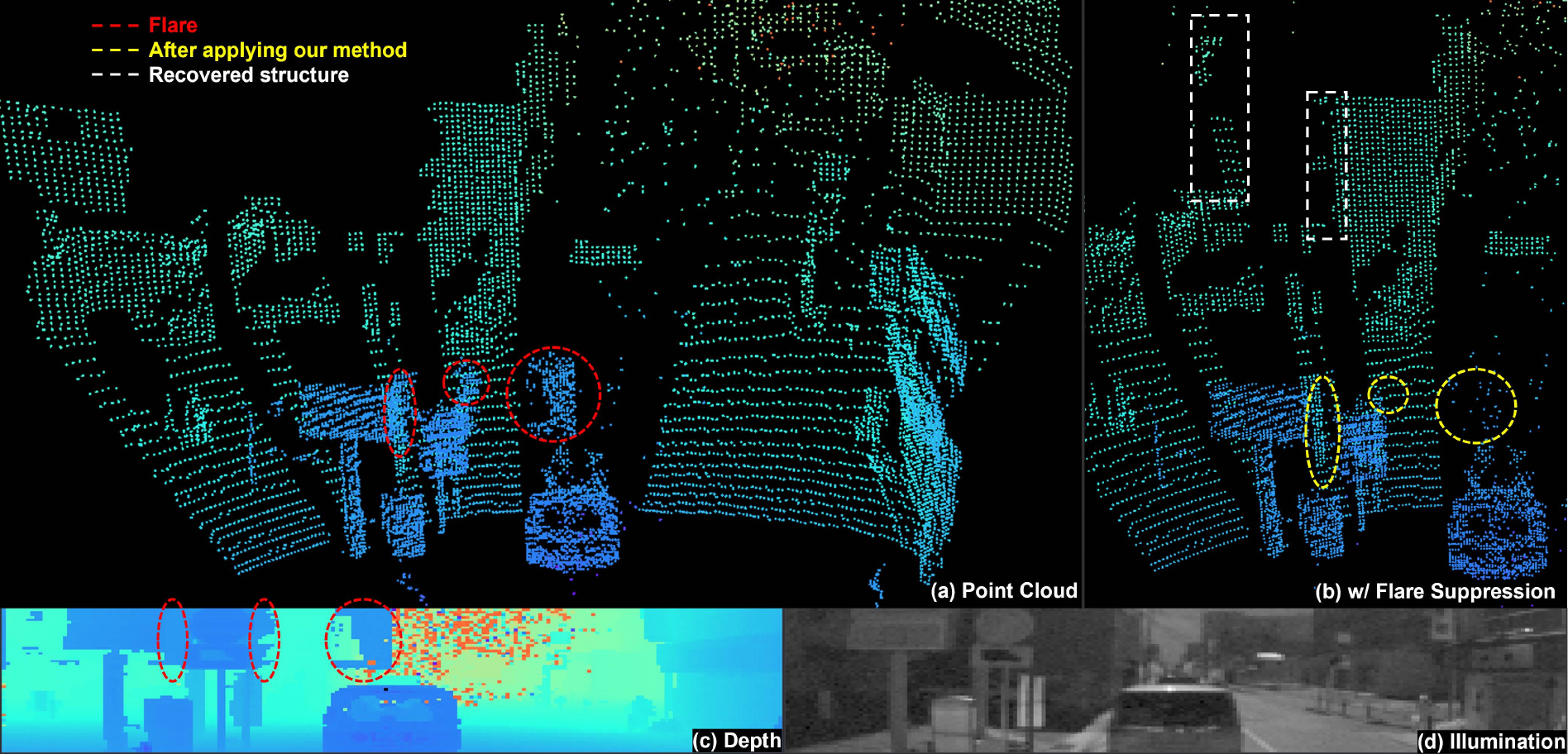}
    \caption{Illustration of SPAD LiDAR flare effects in point clouds. 
    (a) The captured 3D point cloud. 
    (b) The point cloud after flare suppression. 
    (c) The depth map. 
    (d) The ambient illumination image. 
    Excessive photon returns from retro-reflective surfaces, such as traffic signs, can produce strong flare that leads to false or exaggerated structures in the point cloud, potentially distorting geometric measurements. This highlights the importance of flare suppression for safety-critical autonomous systems.}
    \label{fig:pcd}
\end{figure*}

LiDAR has become a cornerstone sensor for autonomous driving and robotics. Among various implementations, SPAD-based LiDAR systems are emerging due to their high sensitivity and precise depth sensing capabilities~\cite{li2020lidar, tashiro2023spad}. However, these systems are highly susceptible to flare caused by excessive photon arrivals or pile-up effects~\cite{ficorella2016crosstalk, jahromi2018timing}, which can distort geometric shapes and produce spurious obstacles, as illustrated in Figure~\ref{fig:pcd}.
Existing methods for suppressing flare in SPAD LiDAR primarily operate at the hardware or signal-processing levels~\cite{rech2008optical, jahromi2018timing}. These non-learnable approaches, while effective in controlled settings, rely on rule-based procedures and are largely dependent on specific system configurations, limiting their generalization across different sensing scenarios. 

To address these limitations, we reformulate flare suppression as a semantic segmentation task, enabling data-driven learning of geometric and photometric cues directly from SPAD measurements.
We first benchmark representative segmentation models on the newly introduced \texttt{FLARE} dataset, revealing that they struggle to accurately distinguish flare from valid structures, particularly along boundaries where flare closely adjoins real objects. 
We attribute this limitation primarily to their inability to leverage the unique multi-echo characteristics of SPAD signals. 
Specifically, in typical processing strategies~\cite{gyongy2021direct, tontini2020numerical}, the first-echo depth and intensity are retained when converting raw SPAD signals into point clouds, while later echoes, which often provide complementary depth information for regions corrupted in the first echo, are discarded, thereby limiting effective discrimination between flare and valid structures.

Building on this insight, we propose \textbf{PILF}, a learning-based approach that directly leverages raw SPAD measurements to exploit their inherent multi-echo and photometric information. 
It treats the first and second echoes, together with ambient illumination, as distinct modalities, aggregating cross-echo information while jointly encoding geometric and photometric features.

Our \textbf{contributions} are threefold. 
\emph{First}, we reformulate SPAD LiDAR flare suppression as a semantic segmentation task. 
\emph{Second}, we design a physically informed multi-echo representation and fusion framework for distinguishing flare from valid structures. 
\emph{Third}, we release \texttt{FLARE}, a SPAD LiDAR dataset covering diverse real-world flare scenes, and provide systematic evaluation against representative segmentation baselines.

\section{Related Work}
\paragraph{SPAD-based LiDAR Flare Suppression.} 
Flare in SPAD-based LiDAR systems primarily arises from excessive photon returns and pile-up effects, which induce optical cross-talk among neighboring SPAD pixels and produce saturated, spatially spread false returns that distort depth estimation and scene geometry.
Traditional flare suppression methods primarily operate at the hardware or signal-processing levels~\cite{villa2021spads,incoronato2021statistical}.
Hardware approaches mitigate detector non-idealities through sensor design and calibration, such as optical isolation, trench structures, on-chip time gating, and advanced readout architectures~\cite{veerappan2014substrate,ficorella2016crosstalk,jahromi2018timing,cusini2022historical}.
Signal-processing approaches suppress distortions from raw SPAD measurements or histograms using techniques such as peak suppression, temporal gating, photon modeling, and asynchronous acquisition~\cite{gupta2019photon,chen2020data,gupta2019asynchronous,po2022adaptive}.
While effective under controlled conditions, these approaches are largely configuration-dependent, offering limited adaptability and lacking a unified framework for integrating physical cues~\cite{chen2020data}.
Recent learning-based waveform methods study related artifacts, such as ghosts~\cite{ikeda2026ghost}, assuming dense temporal signals~\cite{scheuble2025lidar}, whereas we target a practical bandwidth- and latency-constrained setting where full histograms are typically unavailable to downstream perception modules.

\paragraph{Learning-based Semantic Segmentation.}
Recent advances in semantic segmentation have significantly improved performance in both 2D and 3D scene understanding.
In the 2D domain, segmentation has evolved from convolution-based architectures~\cite{long2015fully, chen2018encoder, ronneberger2015u} to transformer-based frameworks~\cite{xie2021segformer, liu2021swin}, achieving strong spatial reasoning and contextual modeling.
However, these methods are typically designed for RGB or single-modality images, lacking access to geometric or temporal cues that are crucial in LiDAR sensing. 
In the 3D domain, segmentation approaches learn spatial semantics across various representations, including point-based~\cite{qi2017pointnet, qi2017pointnet++, wu2024ptv3}, voxel-based~\cite{liu2019point, choy20194d, zhou2020cylinder3d}, and range-view formats~\cite{milioto2019rangenet++, cortinhal2020salsanext, kong2023rethinking, ando2023rangevit}.
While effective at capturing geometric context, these methods generally overlook the physical principles underlying point cloud formation, such as multi-echo photon distributions, which are crucial for distinguishing flare from valid structures. 
This limits their direct applicability to SPAD flare suppression.

\section{Methodology}
\subsection{Physics-Guided Input Representation}
\label{sec:3.1}
\paragraph{Conventional Data Processing.} 
In a typical LiDAR scan, time-resolved SPAD measurements are collected for each laser pulse. Commonly, the first echo is retained and converted into a 3D point cloud~\cite{chen2020data}, which can then be projected onto a 2D cylindrical range view of resolution \(H \times W\), where \(H\) and \(W\) denote the height and width of the range view. 
Each point \((p_x, p_y, p_z)\) is mapped to its corresponding pixel \((i,j)\) as follows:
\begin{equation}
    \left(
    \begin{aligned}
        &i \\
        &j
    \end{aligned}
    \right)
    = 
    \left(
    \begin{array}{c}
        \frac{1}{2} \left[ 1 - \arctan(p_y, p_x) \pi^{-1} \right] W \\
        \left[ 1 - \left( \arcsin(p_z, r^{-1}) + \phi_{down} \right) \xi^{-1} \right] H
    \end{array}
    \right),
    \label{eq:rv}
\end{equation}
where \(r = \sqrt{p_x^2 + p_y^2 + p_z^2}\) denotes the point depth, and \(\xi = |\phi_{down}| + |\phi_{up}|\) represents the vertical field-of-view of the sensor. 
By retaining only the first echo, conventional processing discards subsequent echoes that may contain valuable information related to the flare. This limitation motivates leveraging multi-echo signals to form a richer, multi-modal representation.

\begin{figure}[t]
    \centering
    \begin{minipage}[t]{0.48\linewidth}
        \centering
        \includegraphics[width=\linewidth]{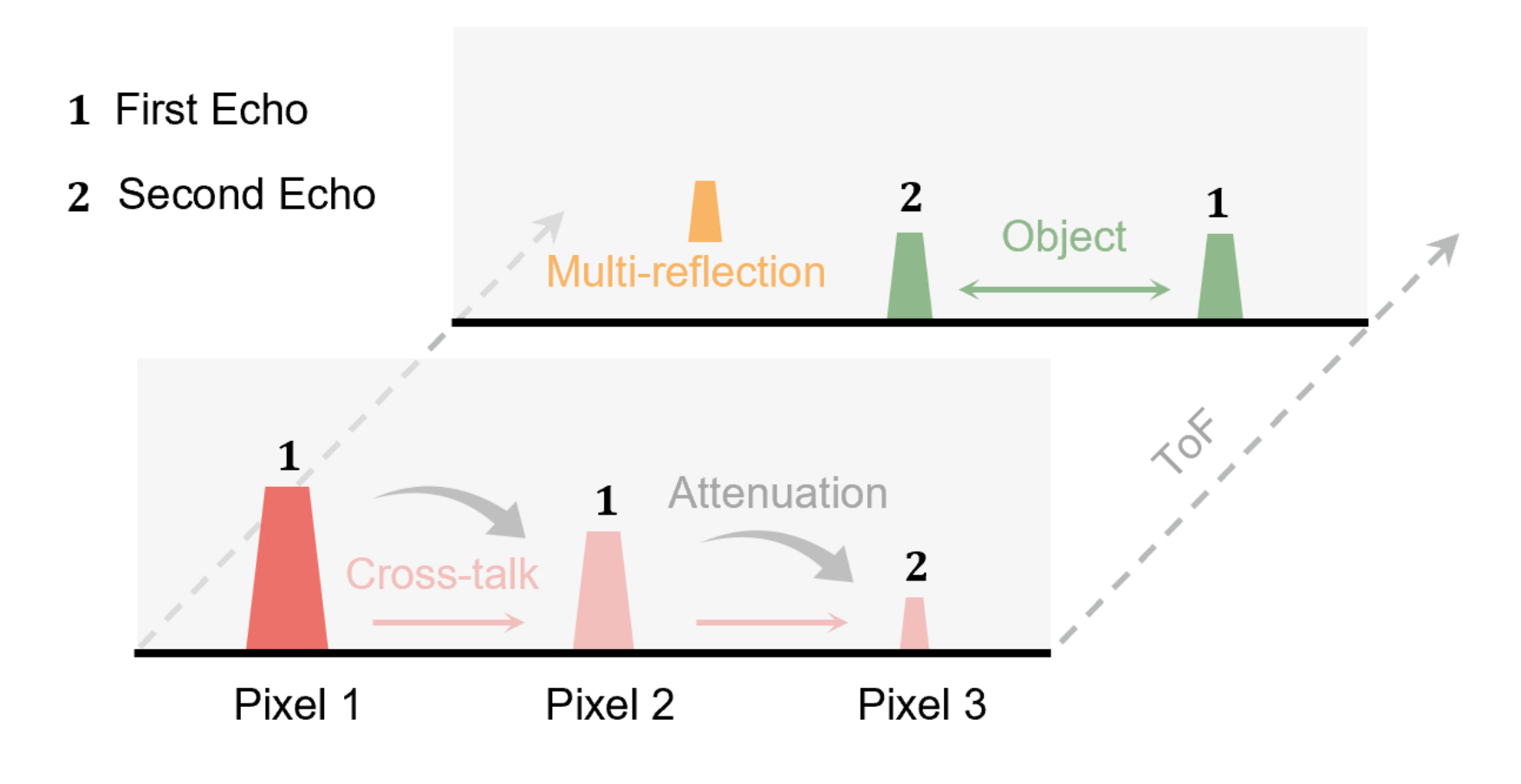}
        \captionof{figure}{Illustration of SPAD flare formation. Excessive photons cause flare to occupy the first echo in neighboring pixels, while valid returns appear in later echoes.}
        \label{fig:form}
    \end{minipage}
    \hfill
    \begin{minipage}[t]{0.48\linewidth}
        \centering
        \includegraphics[width=\linewidth]{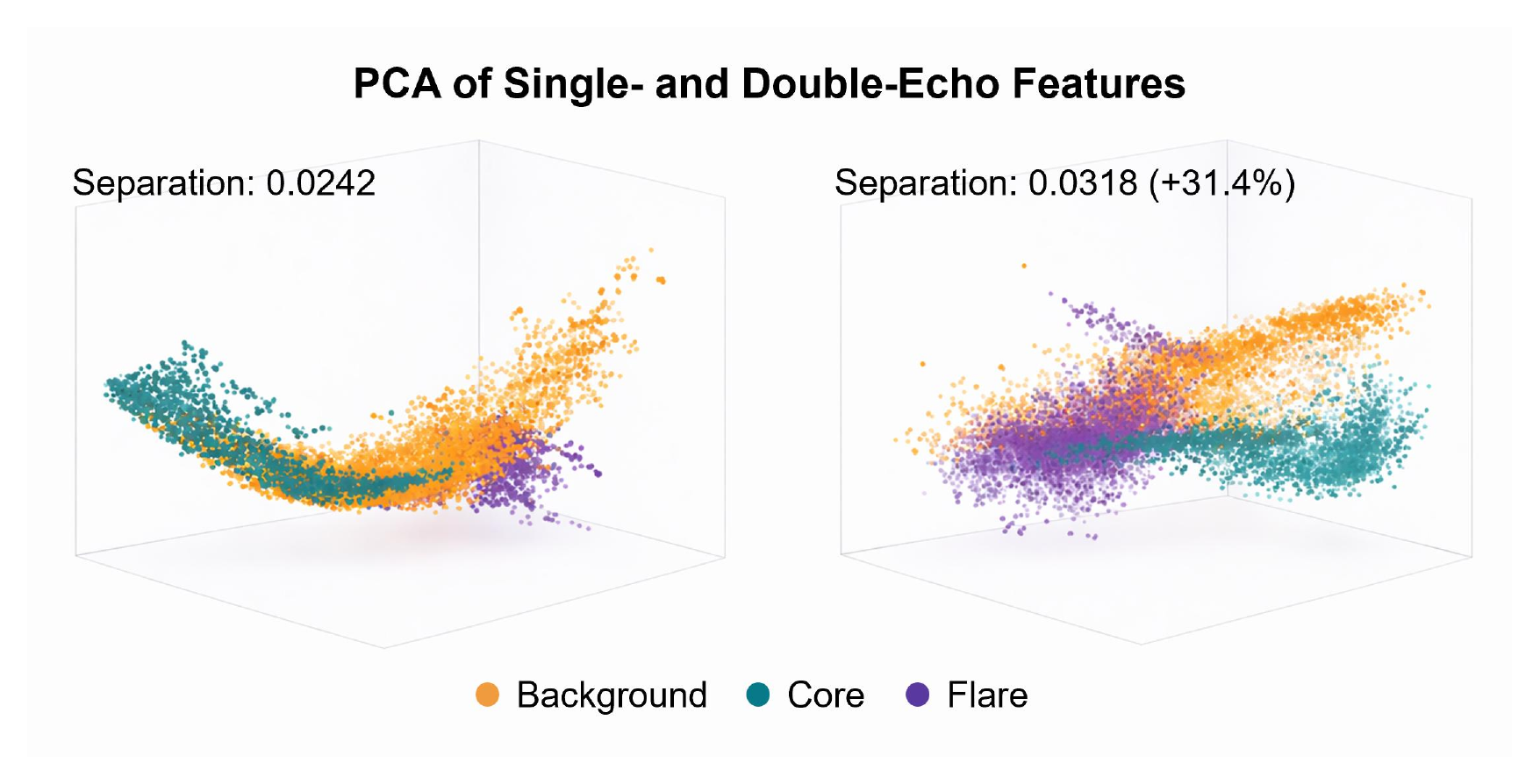}
        \captionof{figure}{PCA-based 3D scatter plots showing up to 10,000 points per class. Utilizing both first and second echoes improves class separability by approximately 30\%.}
        \label{fig:pca}
    \end{minipage}
\end{figure}

\paragraph{Multi-Echo Characteristics.} 
Figure~\ref{fig:form} illustrates that excessive photons from a retro-reflective surface, defined as the core region, are received by SPAD pixel \(1\), inducing optical crosstalk that oversaturates neighboring pixels \(2\). 
In this region, the flare intensity is significantly higher than that of valid object returns, forcing the true signal to appear in the second echo. As the distance from the reflective source increases to pixel \(3\), the flare intensity gradually decays below the object signal, allowing the valid response to reappear in the first echo. 
Excessive returns at pixel \(1\) may also trigger multi-reflections, which generate secondary interference echoes at approximately twice the path length.

These effects corrupt the first-echo measurements, especially near core regions, while later echoes often preserve complementary geometric-photometric cues for distinguishing flare from valid structures.
Figure~\ref{fig:pca} shows PCA-based 3D scatter plots, where using both echoes improves class separability by around 30\%, highlighting the benefit of multi-echo information.

\paragraph{Multi-Modal Input Construction.} 
Based on these observations, we leverage the first two echoes of the SPAD LiDAR signal, each providing three physical measurements: \emph{intensity}, \emph{depth}, and \emph{pulse width}. 
Intensity represents reflected photon counts, depth corresponds to time of flight, and pulse width measures the temporal spread of the returned pulse.
We further approximate flare as an excessive illumination component and introduce an ambient channel to help separate flare responses from intrinsic surface reflectivity~\cite{land1971lightness}.

\begin{figure}[t]
    \centering
    \includegraphics[width=0.95\columnwidth]{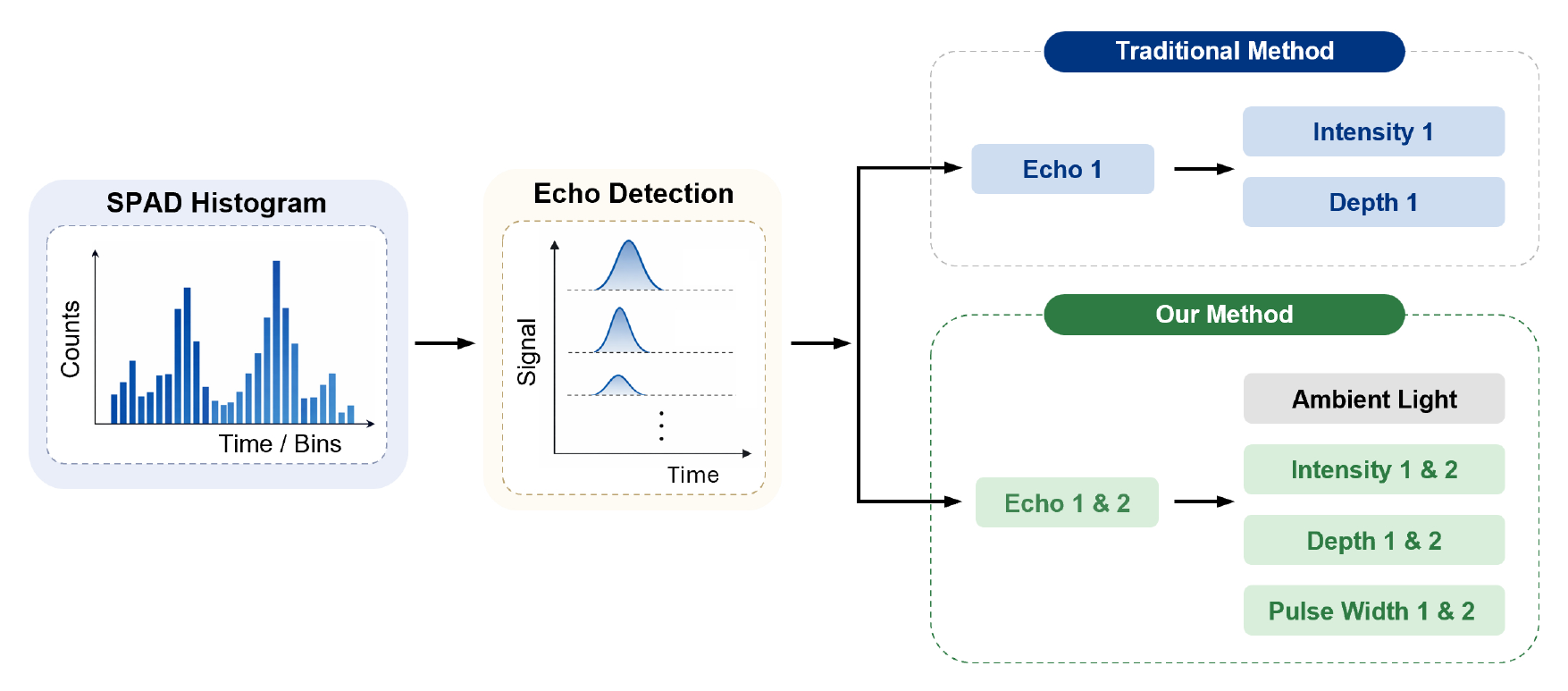}
    \caption{Comparison of conventional and proposed input construction pipelines.
    Traditional processing retains only the first echo.
    Our method selects two echoes and incorporates ambient illumination to construct a 7-channel representation.}
    \label{fig:process}
\end{figure}

Accordingly, our input is a multi-modal 7-channel range-view image.
As illustrated in Figure~\ref{fig:process}, conventional SPAD LiDAR processing often relies on selected first-echo attributes for downstream representation. 
In contrast, our method uses the first two echoes together with ambient illumination and pulse-width information to form a richer multi-modal representation.
Due to the fixed SPAD scanning layout, the measurements naturally form a dense 2D grid aligned with angular coordinates, preserving native spatial resolution without additional resampling.
The resulting representation remains physically interpretable while providing richer signal diversity for learning-based flare suppression.
\subsection{Physically-Informed Flare Segmentation}
\label{sec:3.2}
\begin{figure}[t]
    \centering
    \includegraphics[width=1.0\linewidth]{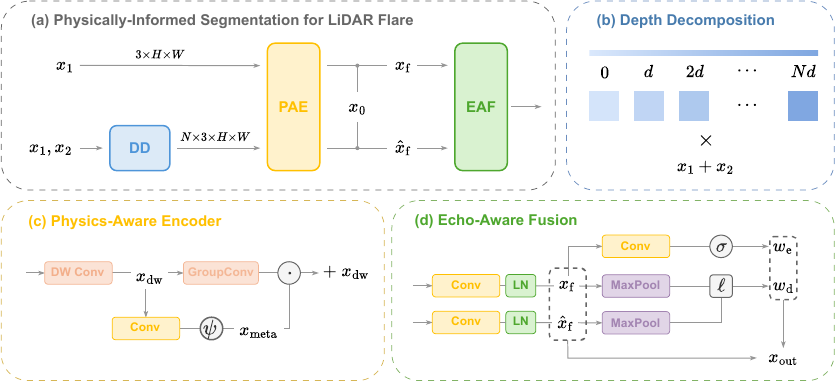}
    \caption{Overview of PILF. 
    (a) The overall architecture processes first echo \(x_1\), second echo \(x_2\), and ambient illumination \(x_0\) through two branches. 
    (b) Depth Decomposition (DD) organizes multi-echo signals into depth-aware bins. 
    (c) Physics-Aware Encoder (PAE) extracts modality-aware geometric-photometric features. 
    (d) Echo-Aware Fusion (EAF) adaptively aggregates echo and illumination cues for segmentation.}
    \label{fig:pilf}
\end{figure}

As shown in Figure~\ref{fig:pilf}, PILF consists of two branches incorporating physical priors. The single-echo branch focuses on the first echo \(x_1\), while the multi-echo branch takes both the first \(x_1\) and second \(x_2\) echoes as input by applying Depth Decomposition (DD). Both branches process the same underlying physical properties and therefore share the Physics-Aware Encoder (PAE), which extracts geometric-photometric representations separately for each modality. With ambient illumination \(x_0\), the resulting features are aggregated by the Echo-Aware Fusion (EAF), which adaptively balances complementary information from the first and second echoes to form a unified representation. 
This representation is then fed into the backbone to predict a mask with three classes: flare, core (its corresponding source), and background. 

\paragraph{Depth Decomposition.}
Multi-echo SPAD signals often contain substantial noise along the depth dimension, which is randomly distributed and lacks spatial continuity, whereas valid measurements are primarily obtained from the first echo.
Fusing such noisy inputs directly complicates model learning. In contrast, flare and its corresponding core exhibit strong spatial continuity within specific depth ranges, and objects partially occluded in the first echo often reappear in subsequent echoes while maintaining consistent depth.

\begin{figure}[t]
    \centering
    \includegraphics[width=0.8\linewidth]{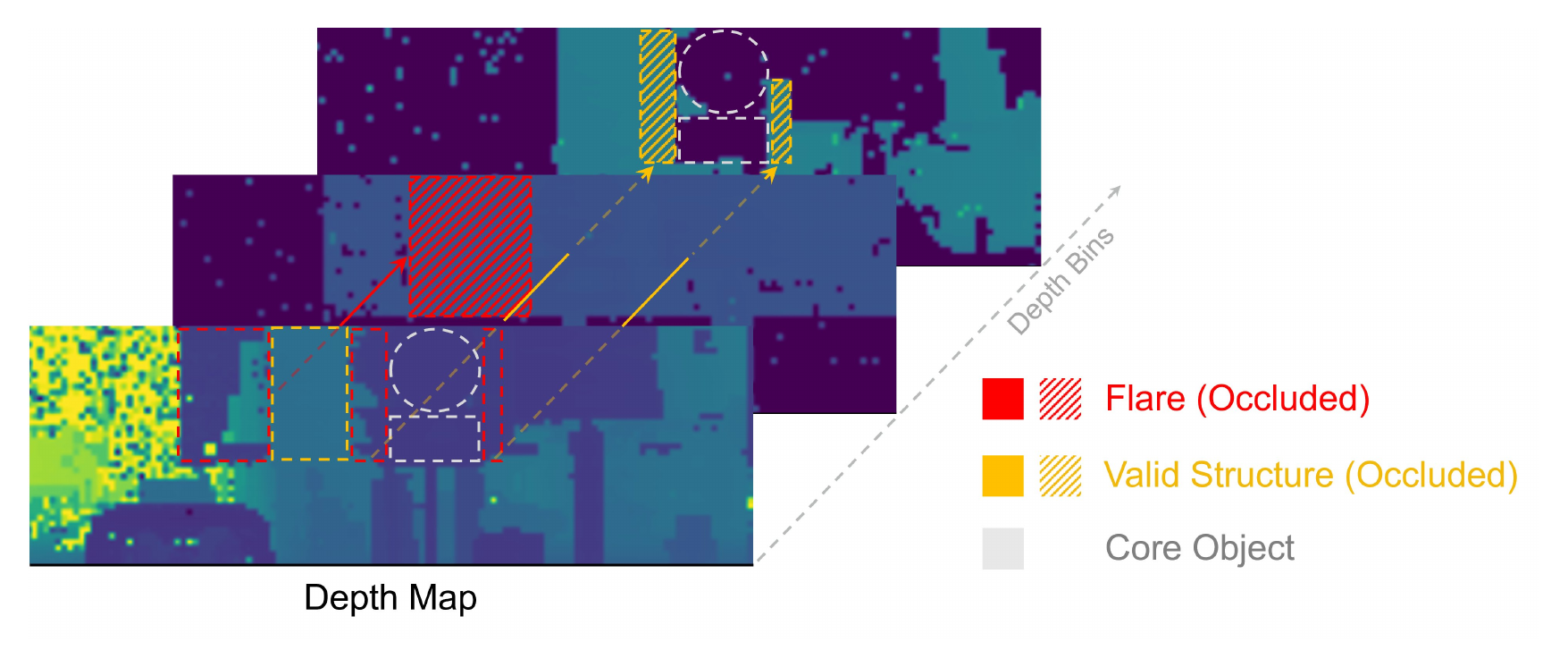}
    \caption {Example of depth decomposition. Within a specific depth range, objects occluded in the second echo, as well as potential flare regions, appear more completely after being merged with the first echo.}
    \label{fig:dd}
\end{figure}

We discretize the depth space into \(N\) bins according to the depth range \(D\) derived from the first echo. This operation not only aggregates spatially coherent flare and object features across echoes but also disperses the randomly distributed noise, as illustrated in Figure~\ref{fig:dd}. 
A binary mask for each bin \(n\) with depth range \(d_n = \left(\frac{\left(n-1\right)\cdot D}{N}, \frac{n\cdot D}{N}\right)\) is defined as:
\begin{equation}
    B^{(n)}(i,j) = 
    \begin{cases}
    1, & \text{if } d(i,j) \in d_n, \\
    0, & \text{otherwise}.
    \end{cases}
\end{equation}
Here, \(d(i,j)\) denotes the depth at pixel \((i,j)\). Within each bin \(n\), masks \(B^{(n)}_1\) and \(B^{(n)}_2\) for the first and second echoes are applied to extract depth-aware features:
\begin{equation}
    \hat{x}^{(n)} = B^{(n)}_1 \odot x_1 +  B^{(n)}_2 \odot x_2, \quad n = 1, \dots, N.
\end{equation}
The resulting tensor \(\hat{x} \in \mathbb{R}^{N \times 3 \times H \times W}\) aggregates features within each depth bin.

\paragraph{Physics-Aware Encoder.}
In our range-view representation, intensity, depth, and pulse width share pixel coordinates but describe different aspects of the photon-return distribution along the same projection ray. 
Their responses are therefore not strictly aligned in 3D: intensity may integrate multiple reflections, depth may be biased by strong returns, and pulse width captures the temporal spread within a depth interval.
Such modality-dependent responses introduce local geometric inconsistencies: neighboring pixels may correspond to different depths or 3D structures, and even the same surface point can produce inconsistent responses across modalities. 
For example, strong retro-reflections may broaden the pulse width or bias the estimated depth. 
Consequently, directly sharing convolutional weights across modalities can cause physical aliasing, entangle modality-specific geometric cues, and produce physically inconsistent features.

To address this, we propose PAE, which learns modality-specific spatial representations while maintaining cross-modal consistency. 
Given a three-modal input \(x\), it first applies depth-wise convolution to extract independent modality-wise features, yielding \(x_{\text{dw}}\). 
A suppressive modulation map is then computed by applying a convolution and the meta function \(\psi(\cdot)\):
\begin{equation}
    x_{\text{meta}} = \psi\left(\underset{3 \times 3}{\text{Conv}}\left(x_{\text{dw}}\right)\right)
\end{equation}
where \(\psi(\cdot)\) mitigates physically unstable responses, such as over-strong reflections or depth outliers. 
Table~\ref{tab:param} compares performance under different \(\psi\) formulations. 
Finally, modality-wise features are updated by a group convolution and combined with the original features through a residual gated transformation:
\begin{equation}
    \text{PAE}(x) = x_{\text{dw}} + \underset{3 \times 3}{\text{GroupConv}}\left(x_{\text{dw}}\right) \odot x_{\text{meta}},
\end{equation}
The outputs are denoted as \(x_{\text{echo}} \in \mathbb{R}^{C\times H\times W}\) for the first echo \(x_1\), and \(\{x_{\text{d}}^{(n)}\}_{n=1}^N\) for the depth-decomposed features \(\hat{x}\), which are stacked for following processing.

\paragraph{Echo-Aware Fusion.}
In flare scenes, multi-modal cues exhibit different reliability and statistics across spatial and depth dimensions, making direct concatenation suboptimal.
First, the large dynamic range between flare and normal regions, together with varying illumination and different valid depth bins, can lead to unstable batch-level feature statistics.
Second, the first echo is reliable in non-flare regions, whereas later echoes provide complementary cues in flare-corrupted areas, making static fusion insufficient.
Third, flare is spatially localized, typically emerging from high-intensity cores and spreading outward, which requires spatially adaptive suppression while preserving valid geometry.

To address these challenges, we propose EAF, which adaptively fuses first-echo and depth-aware features.
Specifically, the illumination map \(x_0\) is first embedded into \(x_{\text{amb}}\), which provides contextual guidance for both streams:
\begin{equation}
\begin{gathered}
    x_{\mathrm{f}} = \mathrm{Concat}(x_{\mathrm{echo}}, x_{\mathrm{amb}}),\\
    \hat{x}_{\mathrm{f}} =
    \left[
    \mathrm{Concat}(x_{\mathrm{d}}^{(1)}, x_{\mathrm{amb}}), \ldots,
    \mathrm{Concat}(x_{\mathrm{d}}^{(N)}, x_{\mathrm{amb}})
    \right].
\end{gathered}
\end{equation}
A convolution followed by LayerNorm is applied to stabilize per-sample feature statistics across samples, which is important for robustly fusing first-echo and depth-bin features.
This operation updates \(x_{\text{f}} \in \mathbb{R}^{C \times H \times W}\) and 
\(\hat{x}_{\text{f}} \in \mathbb{R}^{N \times C \times H \times W}\). 
For simplicity, the following operations are described for each depth bin \(n\). 

To selectively integrate multi-echo features, EAF estimates both a depth-bin coefficient and a spatial gate.
Adaptive max pooling first extracts global descriptors from updated first-echo feature \(x_{\text{f}}\) and each depth-aware feature \(\hat{x}_{\text{f}}^{(n)}\). 
\begin{equation}
    z_{\text{f}} = \text{MaxPool}({x}_{\text{f}}), \quad \hat{z}_{\text{f}}^{(n)} = \text{MaxPool}(\hat{{x}}_{\text{f}}^{(n)}),
\end{equation}

In the depth dimension, these descriptors are combined to compute a bin-wise relevance coefficient, allowing the model to emphasize reliable depth bins.
In the spatial dimension, a gate is generated from the first-echo feature \(x_{\mathrm{f}}\), which carries strong geometric cues, to adaptively modulate each location and reduce over-reliance on corrupted responses.
The two coefficients are computed as:
\begin{equation}
\begin{gathered}
    w_\text{d}^{(n)} = \ell \left( \text{Concat}(z_{\text{f}}, \hat{z}_\text{f}^{(n)})\right),\quad
    w_{\text{e}} = \sigma \left(\underset{3 \times 3}{\text{Conv}} \left({x}_{\text{f}} \right)\right),
\end{gathered}
\end{equation}
where \(\ell(\cdot)\) denotes a learnable projection and \(\sigma(\cdot)\) denotes the sigmoid activation. 

Finally, fusion first aggregates echoes along the depth dimension for depth consistency and then applies spatial modulation for local coherence. 
This approximates the physical intuition of photon integration along depth before projection onto the range-view plane. 
The final representation is:
\begin{equation}
    x_{\text{out}} = x_{\text{f}} + w_{\text{e}} \odot \sum_{n=1}^N \left( w^{(n)}_{\text{d}} ~\hat{{x}}^{(n)}_{\text{f}}  \right),
\end{equation}
where the scalar \(w_{\text{d}}^{(n)}\) is broadcast over the channel and spatial dimensions, and \(\odot\) denotes the Hadamard product.

\paragraph{Backbone Architecture.}
The fusion results are processed by a U-Net~\cite{ronneberger2015u} equipped with a multi-head attention decoder~\cite{vaswani2017attention}. 
Since flare propagation is generally anisotropic and exhibits a dominant spreading direction, we constrain attention to a single principal direction (horizontal in our setup) to better capture this structured pattern. 
This directional restriction enhances long-range feature correlation while maintaining computational efficiency.
\subsection{Mask-Based Flare Suppression}
Given the predicted flare mask, we perform echo-level correction on the structured multi-echo representation parsed from the raw sensor stream.
Each frame contains multiple echo entries per pixel, including temporal and amplitude measurements, while shared metadata such as calibration parameters and acquisition descriptors are preserved during suppression.

Flare predominantly manifests as a premature first photon return caused by multipath or internal reflections, which disrupts the expected echo ordering and corrupts depth consistency.
Let $M$ denote the predicted flare mask. 
For pixels where \(M(i,j)=1\), we locally replace the flare-contaminated first echo with the corresponding second echo:
\begin{equation}
    E'_{i,j,0,:} =
    \begin{cases}
        E_{i,j,1,:}, & M(i,j)=1, \\
        E_{i,j,0,:}, & M(i,j)=0.
    \end{cases}
\end{equation}
After correction, the representation is written back to the original raw data structure while preserving the header and non-echo metadata, ensuring compatibility with downstream processing pipelines.

This strategy modifies only the flare-contaminated primary echo while retaining other metadata and auxiliary echo information.
It is guided by photon arrival ordering in SPAD-based time-of-flight sensing rather than sensor-specific heuristics, allowing integration with hardware-level mitigation techniques.

\section{Experiment}
\subsection{Experimental Setup}
\paragraph{Dataset.}
We evaluate our method on the \texttt{FLARE} dataset, collected using the \texttt{IMX459} SPAD LiDAR.
The dataset covers a wide range of driving and roadside scenes, including urban roads, tunnels, highways, road signs, vehicle rear reflectors, and traffic lights, providing varied flare patterns under different illumination conditions.
All input channels are derived from raw SPAD measurements, without requiring any external imaging sensor.

Each frame is pixel-wise annotated by multiple trained annotators using synchronized intensity, depth, and pulse-width range-view maps.
Flare regions are identified based on depth discontinuities and abnormal intensity patterns, while associated retro-reflective cores are marked as stable high-intensity regions.
A final manual cross-check refines region boundaries, especially around depth transitions, to separate true object surfaces from flare artifacts.
By explicitly labeling both flare and valid core regions, the dataset supports evaluation of flare suppression in terms of both semantic segmentation and structural fidelity.

For training, we use 90\% of the main-road scenes, about 2{,}000 frames. 
The test set is a mixed set containing the remaining 10\% of main-road frames and additional frames from other scenes, about 500 frames in total, to evaluate generalization under unseen conditions.

\paragraph{Normalization.}
We apply modality-specific normalization to the multi-modal input. 
For SPAD LiDAR, the measured intensity \(I\), after scaling to \([0,1)\), reflects photon counts rather than reflected energy \(E\). 
We model its nonlinear response and apply the inverse mapping to approximate the corresponding energy scale:
\begin{equation}
    E = -\frac{1}{\lambda}\log(1 - I).
\end{equation}
Here, \(\lambda\) is a sensor-dependent response coefficient set by the hardware configuration and fixed in our experiments.
The transformed intensity, depth, and pulse-width channels are min--max normalized using first-echo statistics, as both echoes observe the same spatial region with attenuation differences. 
The illumination channel is globally min--max normalized for consistent scaling.

\paragraph{Augmentation.}
The dataset exhibits severe class imbalance, with an approximate pixel ratio of background, flare, and core of $100\!:\!2\!:\!1$. 
To address this, we adopt \emph{FlareAug}, a SPAD-aware augmentation strategy consisting of two operations, crop and fuse, each applied with a probability of $75\%$ during training.

\begin{figure}[t]
    \centering
    \includegraphics[width=1.0\linewidth]{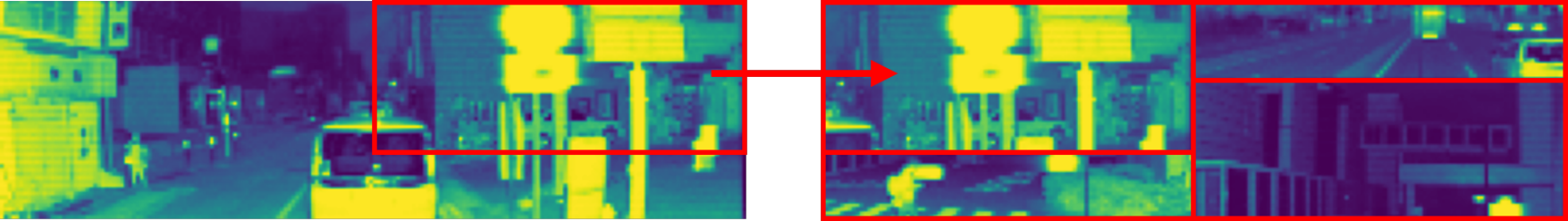}
    \caption{A flare-core region is cropped from the source image and fused into another image while preserving the correspondence between flare and its associated core.}
    \label{fig:aug}
\end{figure}

The crop operation extracts local regions containing both flare and its corresponding core, increasing the frequency and diversity of flare patterns.
The fuse operation resizes the cropped regions to a fixed resolution and inserts them into another training sample in a mosaic-style manner~\cite{byun2024high}, increasing contextual variability, as illustrated in Figure~\ref{fig:aug}.
Based on SPAD array characteristics, flare rarely crosses the image midline and typically extends from its associated core along a dominant horizontal direction in our dataset.
Accordingly, fusion is performed separately on the left and right halves, and each fused flare region is paired with its corresponding core to avoid ambiguous supervision.

\paragraph{Implementation Details.}
PILF is built based on \texttt{MMSegmentation}~\cite{contributors2020mmsegmentation} and trained for \(2\times 10^{4}\) iterations using Adam, with a batch size of \(16\) and a learning rate of \(1\times 10^{-3}\). 
The module configurations are set as follows.
DD uses \(N=16\) depth bins. 
PAE adopts a base channel dimension of \(C=16\), with \(\psi(\cdot)\) set to its best-performing form. 
EAF uses \(C=64\) to match the U-Net backbone, and its bin-wise coefficient \(\ell(\cdot)\) is implemented as a linear layer. 
The network is optimized using an equally weighted combination of cross-entropy and Dice loss~\cite{milletari2016v}.
We evaluate segmentation performance using flare IoU, core IoU, and mean IoU (mIoU), capturing corrupted-region localization, valid-core preservation, and overall segmentation quality, respectively.
\subsection{Flare Segmentation and Suppression}
\paragraph{Segmentation Performance.}
We compare PILF with representative 2D, 3D, and range-view (RV) segmentation methods. 
Here, $(I,D,P)$ denotes the intensity, depth, and pulse width of a single echo with subscripts indicating echo indices, $L$ denotes ambient illumination, and $(x,y,z)$ denotes 3D point coordinates.
For fairness, each method is evaluated with its preferred input representation: 2D models use the proposed 7-channel input, while 3D methods operate on point clouds with intensity, and RV methods use depth maps with their native attributes. 
All methods are trained and evaluated under the same settings.
Table~\ref{tab:comparison} reports overall mIoU for flare and core regios. 
Our method achieves the best performance, providing a reliable mask for subsequent flare suppression.
\begin{table}[t]
\centering
\small
\setlength{\tabcolsep}{7pt}
\begin{threeparttable}
\caption{Quantitative comparison of PILF with 2D, 3D, and RV segmentation methods.
Each baseline follows its native input setting.}
\label{tab:comparison}
\begin{tabular}{lr l ccc}
\toprule
Type & Method & Input & Flare & Core & mIoU \\
\midrule
\multirow{8}{*}{2D}
& FCN~\cite{long2015fully} & $(I_{1,2},D_{1,2},P_{1,2},L)$ & 57.08 & 69.22 & 63.15\\
& DeepLabV3\texttt{+}~\cite{chen2018encoder}
& $(I_{1,2},D_{1,2},P_{1,2},L)$ 
& 67.72 & 76.56 & 72.14\\
& 
& \cellcolor{cellgray}\textcolor{fontgray}{$(I_1,D_1,P_1)$} 
& \cellcolor{cellgray}\textcolor{fontgray}{63.21} & \cellcolor{cellgray}\textcolor{fontgray}{75.34} & \cellcolor{cellgray}\textcolor{fontgray}{69.28}\\
& OCRNet~\cite{yuan2020object}
& $(I_{1,2},D_{1,2},P_{1,2},L)$ 
& 65.69 & 73.48 & 69.59\\
& 
& \cellcolor{cellgray}\textcolor{fontgray}{$(I_1,D_1,P_1)$} 
& \cellcolor{cellgray}\textcolor{fontgray}{63.85} & \cellcolor{cellgray}\textcolor{fontgray}{72.89} & \cellcolor{cellgray}\textcolor{fontgray}{68.37}\\
& KNet~\cite{zhang2021k} & $(I_{1,2},D_{1,2},P_{1,2},L)$ & 57.36 & 68.22 & 62.79\\
& SegFormer~\cite{xie2021segformer} & $(I_{1,2},D_{1,2},P_{1,2},L)$ & 52.26 & 69.92 & 61.09\\
& Mask2Former~\cite{cheng2022masked} & $(I_{1,2},D_{1,2},P_{1,2},L)$ & 52.95 & 69.21 & 61.08\\
\midrule
\multirow{4}{*}{3D}
& PVCNN~\cite{liu2019point} & $(x,y,z,I_1)$ & 35.40 & 42.10 & 38.75\\
& RandLANet~\cite{hu2020randla} & $(x,y,z,I_1)$ & 65.40 & 68.20 & 66.80\\
& Cylinder3D\textsuperscript{*}~\cite{zhou2020cylinder3d} & $(x,y,z,I_1)$ & 70.17 & 72.57 & 71.37\\
& PTv3~\cite{wu2024ptv3} & $(x,y,z,I_1)$ & 57.01 & 54.17 & 55.59\\
\midrule
\multirow{4}{*}{RV}
& RangeNet\texttt{++}~\cite{milioto2019rangenet++} & $(D_1,x,y,z,I_1)$ & 48.83 & 46.69 & 47.76\\
& SalsaNext~\cite{cortinhal2020salsanext} & $(D_1,x,y,z,I_1)$ & 65.66 & 76.74 & 71.20\\
& RangeFormer\textsuperscript{†}~\cite{kong2023rethinking} & -- & -- & -- & --\\
& RangeViT~\cite{ando2023rangevit} & $(D_1,x,y,z,I_1)$ & 56.34 & 57.38 & 56.86\\
\midrule
Ours & PILF & $(I_{1,2},D_{1,2},P_{1,2},L)$ &
\cellcolor{cellgreen}\textcolor{fontgreen}{78.58} &
\cellcolor{cellgreen}\textcolor{fontgreen}{80.06} &
\cellcolor{cellgreen}\textcolor{fontgreen}{79.32}\\
\bottomrule
\end{tabular}
\begin{tablenotes}
\footnotesize
\item[*] Voxel predictions are interpolated to point-level labels for fair comparison.
\item[†] Indicates no official code released.
\end{tablenotes}
\end{threeparttable}
\end{table}

\paragraph{Input Representation.}
To isolate the effect of the proposed input formation, we additionally evaluate OCRNet and DeepLabV3\texttt{+} with first-echo input $(I_1,D_1,P_1)$, shown as gray rows in Table~\ref{tab:comparison}.
Our multi-modal input consistently improves both baselines, confirming the benefit of incorporating second-echo and ambient illumination cues for flare segmentation.

\begin{figure}[t]
    \centering
    \small
    \includegraphics[width=1.0\linewidth]{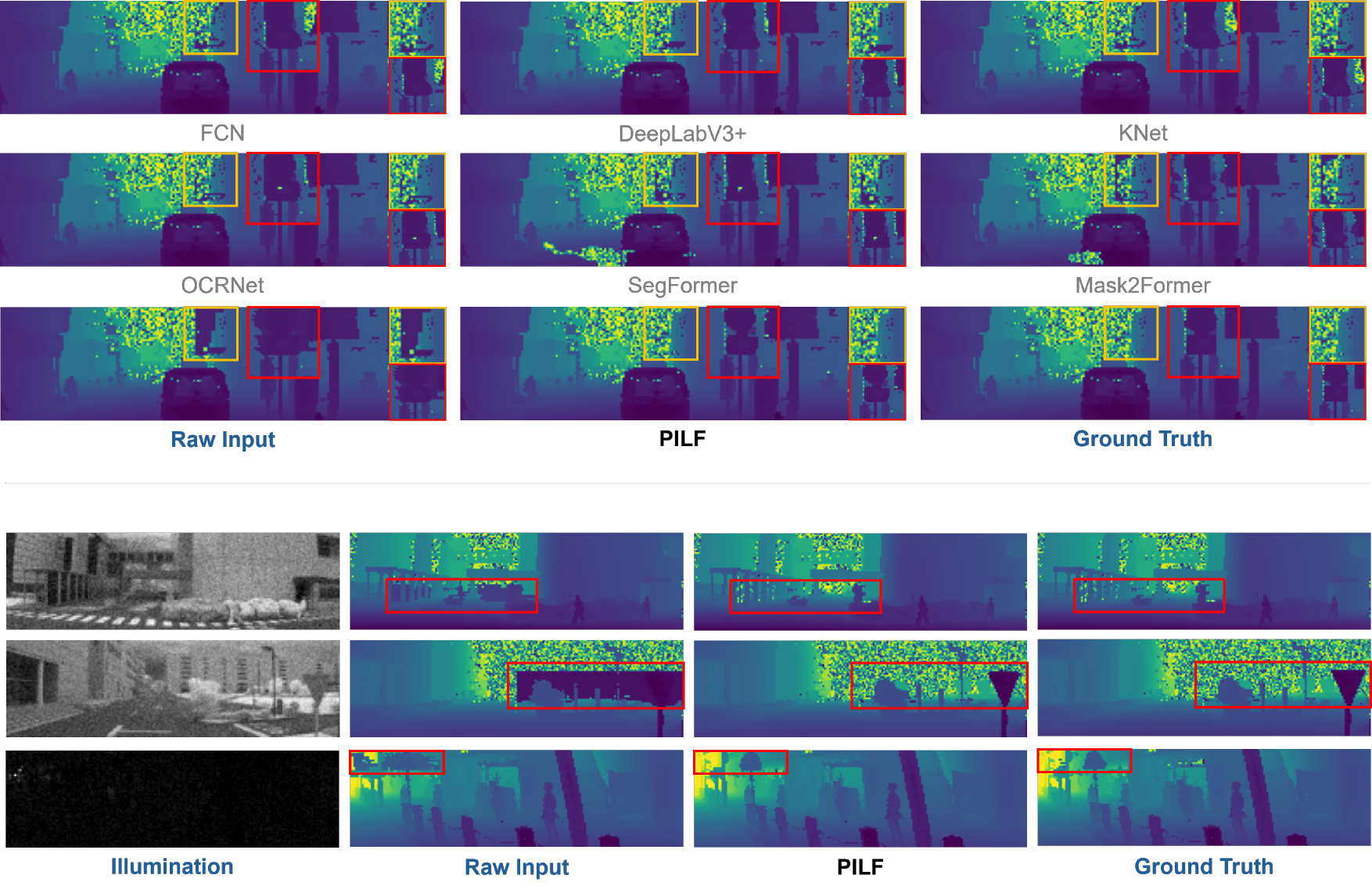}
    \caption{Qualitative visualization of depth maps after flare suppression. 
    The upper block compares different 2D segmentation methods, while the lower block shows additional scenarios under diverse illumination conditions and flare patterns. 
    PILF achieves more complete flare suppression and better structural preservation.}
    \label{fig:vis}
\end{figure}

\paragraph{Suppression Results.}
The predicted mask is used to suppress flare-contaminated echoes, yielding more plausible depth maps and improved scene geometry.
Figure~\ref{fig:vis} presents qualitative suppression results. 
The upper block compares depth maps obtained using segmentation masks from different 2D methods, where PILF more completely removes flare artifacts while preserving valid object structures.
The lower block shows additional scenarios under diverse illumination conditions and flare patterns, demonstrating consistent suppression quality.

\paragraph{Depth Correction.} 
We further evaluate the effect of flare suppression on depth recovery using an annotation-guided echo correction as the reference. 
This reference is obtained by applying the same echo replacement rule with the annotated flare mask. 
We compare the original first-echo depth and the predicted corrected depth against this reference in flare and foreground regions, where foreground denotes the union of flare and core pixels. 
MSE and MAE are computed on normalized depth values, as shown in Table~\ref{tab:d_eval}.
\begin{table}[t]
    \centering
    \small
    \setlength{\tabcolsep}{5pt}
    \caption{Depth-level evaluation before and after flare suppression. Metrics are computed against annotation-guided echo correction. Lower is better.}
    \label{tab:d_eval}
    \begin{tabular}{lcccc}
        \toprule
        \multirow{2}{*}{Setting}
        & \multicolumn{2}{c}{Flare}
        & \multicolumn{2}{c}{Foreground} \\
        \cmidrule(lr){2-3} \cmidrule(lr){4-5}
        & MSE \(\downarrow\) & MAE \(\downarrow\)
        & MSE \(\downarrow\) & MAE \(\downarrow\) \\
        \midrule
        Before 
        & \(0.2275{\scriptstyle\pm0.1535}\)
        & \(0.3849{\scriptstyle\pm0.1601}\)
        & \(0.1447{\scriptstyle\pm0.1014}\)
        & \(0.2447{\scriptstyle\pm0.1165}\) \\
        After 
        & \(0.0321{\scriptstyle\pm0.0398}\)
        & \(0.0613{\scriptstyle\pm0.0608}\)
        & \(0.0219{\scriptstyle\pm0.0211}\)
        & \(0.0413{\scriptstyle\pm0.0341}\) \\
        Reduction
        & \(85.89\%\)
        & \(84.08\%\)
        & \(84.85\%\)
        & \(83.11\%\) \\
        \bottomrule
    \end{tabular}
\end{table}

\paragraph{Practical Discussion.}
Failure cases mainly arise from imperfect flare masks, which may cause under- or over-correction. 
Multiple corrupted echoes or missing reliable later echoes can also limit recovery quality, although these cases are relatively rare in our observations.
By performing suppression directly in the 2D range-view domain, PILF avoids computationally intensive 3D convolutional processing. 
The end-to-end latency is \(37\) ms per frame on a single \texttt{RTX 3090}, which is compatible with \(10\) fps acquisition.
This suggests that deployment on edge devices may be feasible with further optimization.

\subsection{Ablation Study}
\label{sec:ablation}
\paragraph{Hyperparameters.} 
We ablate the number of depth bins \(N\) in DD and the meta function \(\psi(\cdot)\) in PAE.
Each factor is varied independently while keeping the other at its best-performing setting, and each configuration is trained and tested five times.
As shown in Table~\ref{tab:param}, \(N=16\) and \(\psi(x)=1/e^{x^2}\) achieve the best mIoU, while performance remains relatively stable across different settings.

\paragraph{Modalities.}
We ablate the second echo \(x_2\) and ambient illumination \(x_0\) by zeroing out the corresponding inputs while keeping the architecture unchanged.
As shown in the first block of Table~\ref{tab:abl}, removing either modality degrades performance, confirming that multi-echo and ambient cues are complementary for distinguishing flare from valid structures.

\paragraph{Augmentations.} 
We evaluate FlareAug by selectively disabling the crop and fuse operations.
The second block of Table~\ref{tab:abl} shows that both components improve generalization, with their combination achieving the best performance.

\paragraph{Modules.} 
We conduct stepwise ablations to assess each architectural component.
Starting from the full model, we first remove the attention decoder while retaining the U-Net backbone.
We then replace EAF with a shared convolutional layer to preserve channel compatibility.
Finally, we simplify PAE to a single convolutional block and set \(N=1\) to match the expected input dimension from DD.
As shown in the third block of Table~\ref{tab:abl}, each component improves performance, with the complete model achieving the best result.
\begin{table}[t]
    \centering
    \small
    \caption{Ablation study on the number of depth bins \(N\) and meta function \(\psi(\cdot)\).}
    \setlength{\tabcolsep}{7pt}
    \label{tab:param}
    \begin{subtable}[t]{0.38\linewidth}
    \centering
    \begin{tabular}{llc}
    \toprule
    \multicolumn{2}{c}{Depth bins} & mIoU \\
    \midrule
    \multirow{4}{*}{\(N\)}
        & 4 & $78.21{\scriptstyle\pm0.73}$ \\
        & 8 & $77.87{\scriptstyle\pm0.46}$ \\
        & 16 & \cellcolor{cellgreen}\textcolor{fontgreen}{$78.49{\scriptstyle\pm0.29}$} \\
        & 32 & $77.28{\scriptstyle\pm0.60}$ \\
    \bottomrule
    \end{tabular}
    \end{subtable}
    \hfill
    \begin{subtable}[t]{0.58\linewidth}
    \centering
    \begin{tabular}{llc}
    \toprule
    \multicolumn{2}{c}{Meta function} & mIoU \\
    \midrule
    \multirow{4}{*}{\(\psi(\cdot)\)}
        & \(1/(1+x^2)\) & $77.62{\scriptstyle\pm0.30}$ \\
        & \(1/e^{x^2}\) & \cellcolor{cellgreen}\textcolor{fontgreen}{$78.49{\scriptstyle\pm0.29}$} \\
        & \(\cos\left(\pi/2 \cdot \tanh(x)\right)\) & $77.67{\scriptstyle\pm0.41}$ \\
        & \(1/(1+\left|x\right|)\) & $77.83{\scriptstyle\pm0.22}$ \\
    \bottomrule
    \end{tabular}
    \end{subtable}
\end{table}
\begin{table}[t]
    \centering
    \small
    \caption{Ablation studies on input modalities, augmentation operations, and architectural components. All results are averaged over five runs.}
    \setlength{\tabcolsep}{3pt}
    \label{tab:abl}
    \begin{minipage}{0.28\linewidth}
        \centering
        \begin{tabular}{ccc}
            \toprule
            \(x_2\) & \(x_0\) & mIoU \\
            \midrule
            & & \(75.18{\scriptstyle\pm0.45}\) \\
            \checkmark & & \(75.56{\scriptstyle\pm0.81}\) \\
            & \checkmark & \(77.67{\scriptstyle\pm0.22}\) \\
            \checkmark & \checkmark & \cellcolor{cellgreen}\textcolor{fontgreen}{$78.49{\scriptstyle\pm0.29}$} \\
            \bottomrule
        \end{tabular}
    \end{minipage}
    \hfill
    \begin{minipage}{0.28\linewidth}
        \centering
        \begin{tabular}{ccc}
            \toprule
            Crop & Fuse & mIoU \\
            \midrule
            & & \(74.91{\scriptstyle\pm1.30}\) \\
            \checkmark & & \(77.41{\scriptstyle\pm0.53}\) \\
            & \checkmark & \(75.58{\scriptstyle\pm0.56}\) \\
            \checkmark & \checkmark & \cellcolor{cellgreen}\textcolor{fontgreen}{$78.49{\scriptstyle\pm0.29}$} \\
            \bottomrule
        \end{tabular}
    \end{minipage}
    \hfill
    \begin{minipage}{0.42\linewidth}
        \centering
        \begin{tabular}{cccc}
            \toprule
            PAE & EAF & Attn. & mIoU \\
            \midrule
            & & & \(64.41{\scriptstyle\pm0.35}\) \\
            \checkmark & & & \(71.01{\scriptstyle\pm0.49}\) \\
            \checkmark & \checkmark & & \(75.40{\scriptstyle\pm0.54}\) \\
            \checkmark & \checkmark & \checkmark & \cellcolor{cellgreen}\textcolor{fontgreen}{$78.49{\scriptstyle\pm0.29}$} \\
            \bottomrule
        \end{tabular}
    \end{minipage}
\end{table}

\section{Conclusion}
In this paper, we present PILF, a physically informed segmentation method for flare suppression in SPAD-based LiDAR systems.
By reformulating flare suppression as a semantic segmentation problem, PILF learns geometric-photometric cues from raw SPAD measurements while leveraging multi-echo and ambient illumination information. 
This design enables effective discrimination between spurious flare artifacts and valid object structures.
Extensive experiments show that PILF effectively suppresses flare while preserving the structural fidelity of true object surfaces across diverse real-world scenarios.
Operating in the 2D range-view domain, PILF avoids computationally intensive 3D convolutional processing.
Moreover, since suppression is performed at the output level, it does not alter the underlying hardware or signal acquisition process, making it complementary to hardware- and signal-level mitigation strategies and compatible with existing pipelines.
Overall, this work demonstrates the potential of combining physically grounded modeling with learning-based segmentation for mitigating challenging flare artifacts in SPAD-based LiDAR systems.


\section*{Acknowledgements}
We would like to thank Hayakawa Michio from Sony Semiconductor Solutions for his valuable assistance in data collection.

%
%
\bibliographystyle{splncs04}
\bibliography{main}
\end{document}